\PassOptionsToPackage{table}{xcolor}
\documentclass[runningheads]{llncs}

\usepackage{eccv}

\usepackage{eccvabbrv}
\usepackage{graphicx}
\usepackage{booktabs}
\usepackage{wrapfig}
\usepackage{algorithm}
\usepackage{algorithmic}
\usepackage{makecell}
\usepackage{amssymb}
\usepackage{amsmath}
\usepackage{multirow}
\usepackage{makecell}
\usepackage{booktabs}
\usepackage{bm} 
\usepackage{xcolor}
\usepackage{pifont}
\newcommand{\cmark}{\ding{51}}  % ✓
\newcommand{\xmark}{\ding{55}}  % ✗
\definecolor{first}{RGB}{255,200,200}
\definecolor{second}{RGB}{200,255,200}
\definecolor{third}{RGB}{200,200,255}

\usepackage[accsupp]{axessibility}  % Improves PDF readability for those with disabilities.

\usepackage{hyperref}

\usepackage{orcidlink}

\begin{document}

% ---------------------------------------------------------------
% TODO REVIEW: Replace with your title
\title{Not All Pixels Are Equal: Confidence-Guided Attention for Feature Matching} 

% TODO REVIEW: If the paper title is too long for the running head, you can set
% an abbreviated paper title here. If not, comment out.
\titlerunning{Abbreviated paper title}

% TODO FINAL: Replace with your author list. 
% Include the authors' OCRID for the camera-ready version, if at all possible.
% \author{Dongyue Li \and
%  \and
% Third Author\inst{3}\orcidlink{2222--3333-4444-5555}}
\author{Dongyue Li}

% TODO FINAL: Replace with an abbreviated list of authors.
\authorrunning{F.~Author et al.}
% First names are abbreviated in the running head.
% If there are more than two authors, 'et al.' is used.

% TODO FINAL: Replace with your institution list.
% \institute{Princeton University, Princeton NJ 08544, USA \and
% Springer Heidelberg, Tiergartenstr.~17, 69121 Heidelberg, Germany
% \email{lncs@springer.com}\\
% \url{http://www.springer.com/gp/computer-science/lncs} \and
% ABC Institute, Rupert-Karls-University Heidelberg, Heidelberg, Germany\\
% \email{\{abc,lncs\}@uni-heidelberg.de}}

\maketitle

\begin{abstract}
  Semi-dense feature matching methods have been significantly advanced by leveraging attention mechanisms to extract discriminative descriptors. However, most existing approaches treat all pixels equally during attention computations, which can potentially introduce noise and redundancy from irrelevant regions. To address this issue, we propose a confidence-guided attention that adaptively prunes attention weights for each pixel based on precomputed matching confidence maps. These maps are generated by evaluating the mutual similarity between feature pairs extracted from the backbone, where high confidence indicates a high potential for matching. Then the attention is refined through two steps: (1) a confidence-guided bias is introduced to adaptively adjust the attention distributions for each query pixel, avoiding irrelevant interactions between non-overlap pixels; (2) the corresponding confidence map is additionally employed to rescale value features during feature aggregation, attenuating the influence of uncertain regions. Moreover, a classification loss is introduced to encourage the backbone's features to discriminate between matchable and non-matchable regions. Extensive experiments on three benchmarks demonstrate that the proposal outperforms existing state-of-the-art methods.
\end{abstract}

\section{Introduction}
\label{sec:intro}
Local feature matching, which aims to establish point-to-point correspondences between a pair of images, is the prerequisite for a variety of downstream 3D computer vision tasks, including 3D reconstruction \cite{3D, Fast3R, dust3r, splatter}, visual localization \cite{VLOC, tcsvt-vloc1, tcsvt-vloc2, tcsvt-vloc3}, structure from motion (SfM) \cite{SFM, lee2025densesfm, pataki2025mpsfm, he2024dfsfm, light3r}, simultaneous localization and mapping (SLAM) \cite{SLAM, 25slam1, 25slam2, 2025slam3}, \emph{etc}. Despite its significance, local feature matching still remains challenging due to extreme appearance variations caused by viewpoint change, illumination variation, and motion blur.

Presently, most feature matching methods can be divided into 3 categories: sparse \cite{Superglue, lightglue}, semi-dense \cite{LoFTR, Eloftr, aspanformer}, and dense methods \cite{RoMa, dkm}. Sparse methods typically begin by employing an explicit keypoint detector, followed by matching of the detected keypoints. However, in scenes dominated by repetitive patterns, such as indoor environments, these keypoints often exhibit low repeatability, making it difficult for such methods to establish a sufficient number of reliable correspondences. In contrast, dense methods aim to establish correspondences across all pixels, and while they often achieve impressive performance, their high computational cost renders them impractical for real-time applications. Semi-dense methods, which lie between sparse and dense approaches, often adopt a coarse-to-fine paradigm for building matches with sub-pixel accuracy. This kind of method represents a compromise between matching accuracy and computational efficiency.
\begin{figure*}[t]
\centering
\includegraphics[width=\linewidth]{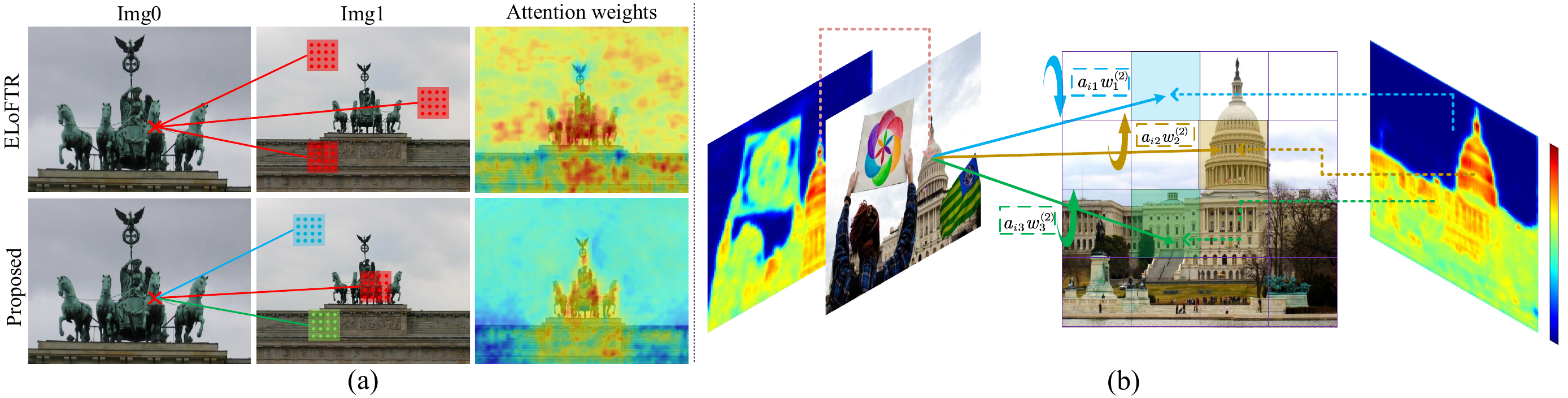}
\vspace{-0.3in}
\caption{(a) Motivation of our method. Red indicates high attention concentration, while green and blue denote low attention.
(b) Graphical illustration of the proposed confidence-guided attention. The left and right images show confidence maps. The color bar indicates pixel-wise matching confidence (high to low).}
\vspace{-0.3in}
\label{motivation}
\end{figure*}

Recent semi-dense matching methods have increasingly relied on attention mechanisms \cite{flashattention} to capture long-range feature dependencies and integrate global contextual cues. This formulation allows the network to produce more discriminative feature representations, resulting in improved robustness and accuracy. LoFTR~\cite{LoFTR}, a pioneering semi-dense matching framework, employs linear attention to reduce the computational cost by lowering the resolution of features at the coarse level. However, it still processes the entire set of coarse-level features, and the linear approximation tends to degrade matching accuracy. To further mitigate computational overhead, ELoFTR \cite{Eloftr} downsamples coarse-level features prior to applying vanilla attention and subsequently upsamples the refined representations. This U-Net-like architecture achieves simultaneous improvements in accuracy and efficiency.
Despite these advances, attention weights of these methods are typically learned without prior knowledge about which pixels are truly informative for matching. Therefore, previous semi-dense methods could assign excessive attention to ambiguous or uninformative regions, such as non-co-visible areas, leading to redundant computations and suboptimal matching accuracy. As illustrated in Fig. \ref{motivation} (a), ELoFTR \cite{Eloftr} tends to attend broadly across the image, including regions that may not contribute to reliable matching. This over-attention introduces additional noise during feature aggregation within the attention mechanism, thereby degrading the representational quality of the learned features.

\begin{figure*}[t]
\centering
\includegraphics[width=\linewidth]{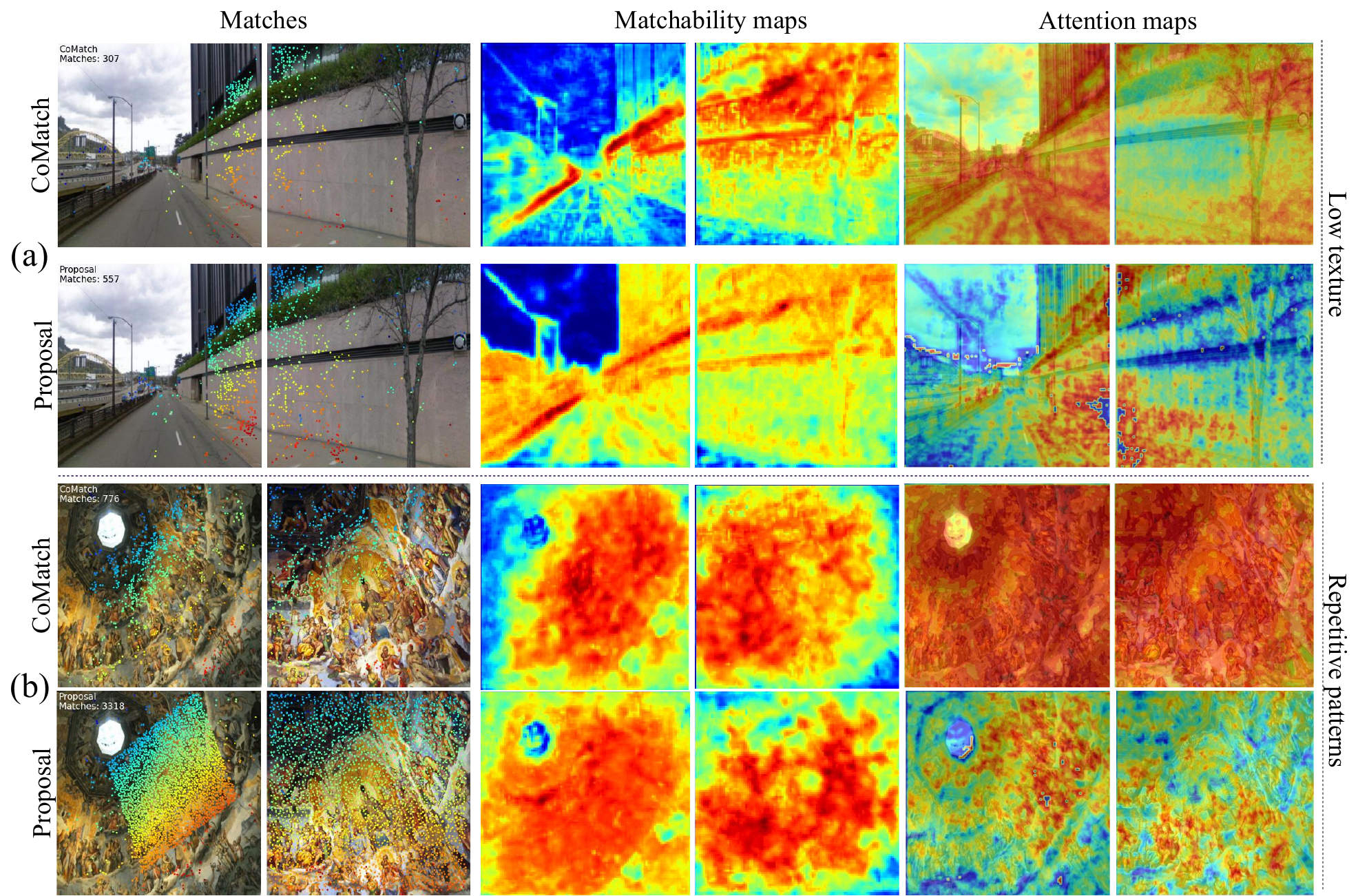}
\vspace{-0.2in}
\caption{Qualitative comparison of CoMatch \cite{comatcher} and ours on images with low texture and repetitive patterns. Correspondences are marked as the same color.}
\vspace{-0.4in}
\label{comatch}
\end{figure*}

To address this issue, several follow-up works have been proposed. ASpanFormer~\cite{aspanformer} adjusts the attention span according to the learned flow maps and performs attention only within these adaptive spans. Although it achieves performance gains, when the estimated flow maps are not sufficiently robust, this localized attention fails to aggregate global context, potentially misleading the matching process.
Concurrent with our work,
CoMatch \cite{comatcher} improves performance by rescaling features based on co-visible scores. However, this design does not alter the inherent attention distribution. Therefore, as shown in Fig.~\ref{comatch}, CoMatch \cite{comatcher} still exhibits two limitations: it fails to further discriminate co-visible pixels with similar appearance in repetitive regions, and it remains susceptible to attention diffusion when confidence maps are unreliable. Our method introduces co-visible priors and an additional learnable parameter to sharpen attention for ensuring that each query pixel only focuses on most similar target ones.

In this paper, we argue that not all pixels contribute equally to the matching process.
We introduce a confidence-guided attention mechanism that explicitly incorporates matching priors. Specifically, we estimate pixel-wise matching confidence by computing a pairwise similarity matrix between dense feature maps of an image pair, followed by maximization to emphasize regions with strong mutual responses. These confidence maps, refined under the supervision of a binary classification loss, serve as spatial priors indicating the significance of each region. As shown in Fig. \ref{motivation} (b), to effectively integrate this prior, we propose the confidence-guided attention. In the first stage, an confidence-guided attention bias is introduced, which effectively sharpens the attention distribution. This differentiable mechanism is regarded as a soft approximation to hard selection, allowing dynamic control over attention sharpness. In the second stage, the value features are adaptively scaled by another confidence map, enabling spatially aware modulation of pixel contributions. This design allows the attention to dynamically suppress unreliable regions and enhance discriminative aggregation, leading to more robust and accurate matching under challenging visual conditions.
To summarize, the main contributions of this work are as follows:
\begin{itemize}
\item  Pixel-wise matching confidence maps are introduced as learnable spatial priors, enabling the network to estimate the reliability of each region.

\item A confidence-guided attention is introduced for refining attention weights at both pre- and post-softmax stages.

\item  Extensive experiments on various benchmarks show that our method outperforms sparse and semi-dense feature matching baselines by a large margin.
\end{itemize}

\section{Related Works}
\textbf{Semi-dense Feature Matching.} 
Attention-based architectures have become the dominant paradigm for semi-dense matching methods. LoFTR \cite{LoFTR} adopts a coarse-to-fine strategy to predict match proposals. MatchFormer \cite{matchformer} proposes an extract-and-match pipeline, where a pure Transformer is utilized to perform feature extraction and matching simultaneously. To enable fine-grained local interactions among pixel tokens, ASpanFormer \cite{aspanformer} is proposed. After each cross-attention phase, flow maps are regressed to adjust the attention span adaptively based on uncertainty prediction. ContextMatcher \cite{contextmatcher} proposes an additional cross-scale Transformer to capture both coarse-grained and fine-grained features directly from the original images. AdaMatcher \cite{adamatcher} is the first to incorporate overlapping knowledge into semi-dense matchers, adopting high-level queries to update feature descriptors and extending beyond one-to-one matching. TopicFM \cite{topicfm} introduces a set of $K$ topics to model images, thereby providing a more interpretable matching process. 
To achieve a manageable computation cost, ELoFTR \cite{Eloftr} adopts a UNet-like structure for the feature description. However, despite the effectiveness of these methods, they still uniformly attend to all pixels, introducing interference from irrelevant regions.

\begin{figure*}[t]
\centering
\includegraphics[width=\linewidth]{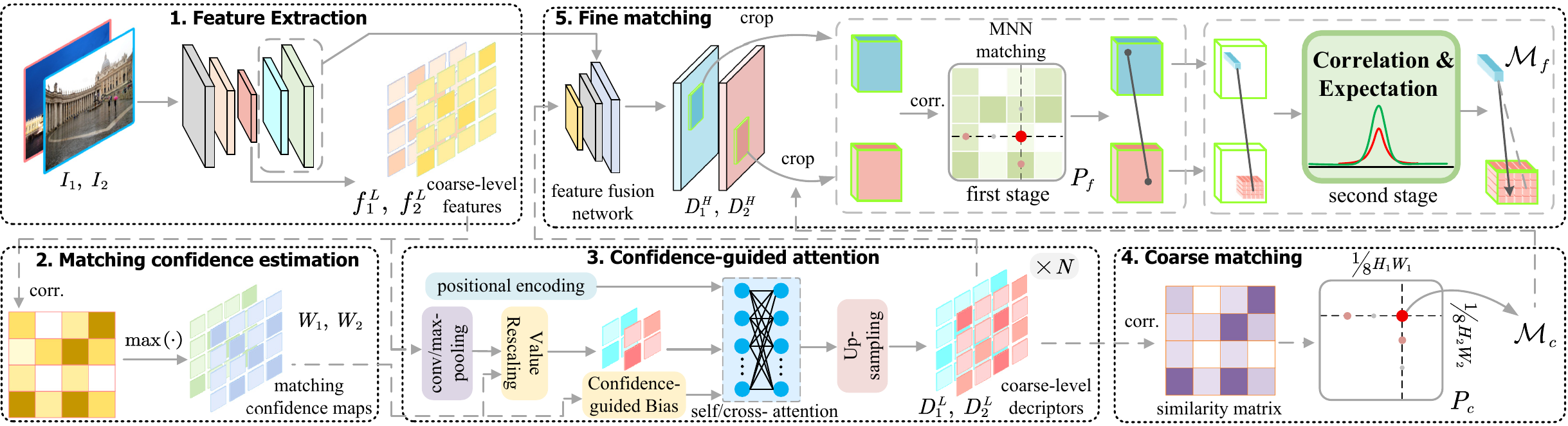} % Reduce the figure size so that it is slightly narrower than the column.
\vspace{-0.25in}
\caption{The overview of the proposed matching pipeline. Matching confidence maps $W_1, W_2$ are estimated through the correlation of coarse-level features from the CNN backbone. These maps are then fed into the confidence-guided attention module to update the coarse-level features for obtaining coarse-level descriptors $D^L_1, D^L_2$. The coarse matches $\mathcal{M}_c$ are then obtained by calculating the similarity of $D^L_1, D^L_2$, followed by the mutual nearest neighbor searching. To refine $\mathcal{M}_c$ to be fine-level, a two-stage refinement is performed to obtain the final matches $\mathcal{M}_f$.}
\vspace{-0.25in}
\label{fig:overview}
\end{figure*}

\noindent\textbf{Scene Aware feature Matching.} Since OETR \cite{OETR} has employed a DETR-like architecture to detect overlapping areas between a pair of images, numerous approaches have tried to leverage the information about overlapping areas to improve the matching results.
SAM \cite{Lu_ICCV} introduces a grouping mechanism to divide all keypoints into two clusters and design a loss function to supervise the grouping module. Scale-net \cite{Scale-net} first proposes the co-visible matching module, which uses the estimated co-visible maps for filtering the matching matrix. MESA \cite{MESA} introduces the additional semantic segmentation phase to segment images for enhancing the performance of existing matchers.
OAMtcher \cite{OAMatcher} directly leverages co-visible masks to segment features and perform attention between segmented features. Although these methods have improved performance, they mainly modify input features without considering the inherent attention distributions, whereas our proposed method provides a more comprehensive way to enhance attention.

\section{Methodology}

An overview of the proposal is presented in Fig. \ref{fig:overview}. A CNN backbone extracts multi-scale image features. Coarse-level features are used to establish initial correspondences, while fine-level features are employed to refine them into final matches. Confidence maps, which indicate the likelihood of each pixel finding a reliable match, are estimated from feature similarities and supervised using a binary classification loss. 
The confidence-guided attention sharpens the attention distribution for high-confidence queries and rescales values through their corresponding confidence.
In the following subsections, we will elaborate on the details of each block as well as the underlying insights.

\subsection{Feature Extraction}
A lightweight backbone with reparameterization \cite{Eloftr} is adopted as the backbone to extract multi-scale feature maps from the given images $I_1 \in \mathbb{R}^{C \times H \times W}$ and $I_2 \in \mathbb{R}^{C \times H \times W}$. $H$ and $W$ represent the height and width of the image, respectively. $C$ denotes the number of channels. Specifically, feature maps from the last layer of the backbone are extracted as coarse-level features, which are denoted as $f^L_1 \in \mathbb{R}^{C_d \times H_c \times W_c}$ and $f^L_2 \in \mathbb{R}^{C_d \times H_c \times W_c}$ ($H_c = H/8, W_c = W/8, C_d = 256$).
In addition to the coarse feature maps, fine feature maps $f^H_1 \in \mathbb{R}^{C_f \times H_f \times W_f}$ and $f^H_2 \in \mathbb{R}^{C_f \times H_f \times W_f}$ ($H_f = H/2, W_f = W/2, C_f = 128$) are also extracted from the shallower layer of the backbone.

\subsection{Matching Confidence Estimation}
% Our key idea is to derive pixel-wise confidence priors by identifying mutually similar regions between image pairs. These priors provide the initial weighting over feature locations, effectively narrowing the attention space to more reliable regions. As a result, the attention is guided to focus on discriminative areas, allowing the final attention weights to be learned more effectively.
Given coarse-level features $f^L_1, f^L_2$, the correlation matrix $S \in \mathbb{R}^{H_cW_c \times H_cW_c}$ is first computed by $S = \frac{1}{\gamma} \left\langle f^L_1,f^L_2\right\rangle$,
where $\left\langle \cdot \right\rangle$ is the dot product. $\gamma$ is the temperature parameter to control the magnitude of $S$. Inspired by the fact that a matchable pixel is likely to have a clear correspondence in the other image, its maximum response score with respect to all candidate points is expected to be higher than that of non-matchable pixels, which lack reliable counterparts. Therefore, the confidence maps are obtained through taking the max value along the column or row dimension, respectively, as $\tilde{W}_1 = \max\limits_{j} S_{ij}, \tilde{W}_2 = \max\limits_{i} S_{ij}$. These maps are subsequently refined to highlight high-confidence regions and suppress ambiguous regions,
\begin{equation}
\hat{W}_m = \sigma(\tilde{W}_m - \tilde{\mathbf{1}} \cdot \text{mean}(\tilde{W}_m)),\\
\end{equation}
where $m$ = 1, 2 and $\tilde{\mathbf{1}}$ denotes a matrix of ones with the same shape as $\tilde{W}_1, \tilde{W}_2$. $\mathrm{mean}(\cdot)$ is to calculate the mean value of all elements and $\sigma(\cdot)$ is the sigmoid function. The channel dimension of $\hat{W}_1, \hat{W}_2$ is 1, forming $\hat{W}_1, \hat{W}_2 \in \mathbb{R}^{H_cW_c \times 1}$.

\subsection{Confidence-guided Attention}

Previous semi-dense methods \cite{ECOTR, LoFTR} compute attention without any priors, inevitably incorporating irrelevant regions. This results in redundant computations and the aggregation of uninformative features, which may hinder performance. To address this issue, we propose a confidence-guided attention mechanism, as illustrated in Fig. \ref{motivation} (b).

For clarity, we present the cross-attention mechanism here, as the self-attention mechanism follows an identical formulation, differing solely in the use of identical inputs. Assuming that input features are $f^L_1, f^L_2, \hat{W}_1, \hat{W}_2$, the $Q, K, V$ can be obtained as follows,
\begin{equation}
\begin{aligned}
&\tilde{Q} = R(\text{Conv2D}(f^L_1)), \tilde{K} = R(\text{MaxPool}(f^L_2)),\\
&\tilde{V} = \text{MaxPool}(f^L_2), W_n = \text{FL}(\text{MaxPool}(\hat{W}_n)),\\
& Q = \text{FL}(\tilde{Q})U_Q, K = \text{FL}(\tilde{K})U_K, V = \text{FL}(\tilde{V})U_V,
\end{aligned}
\end{equation}
where $n$ = 1, 2 and $R(\cdot)$ denotes the RoPE encoding \cite{RoPE}. $\text{Conv2D}(\cdot)$ stands for the convolution for down-sampling, $\text{MaxPool}(\cdot)$ is the max-pooling. FL($\cdot$) represents flattening 2D features to tokens and $U_Q$, $U_K$, $U_V$ are learnable parameters. At this stage, we have $Q, K, V \in \mathbb{R}^{N \times C_d}$ and $W_1, W_2 \in \mathbb{R}^{N \times 1}$, where $N$ is the number of tokens.

\noindent\textbf{Confidence-guided Bias}. Then, we begin by introducing a confidence-guided bias term $B$ into the computation of the attention score $A'$,
\begin{equation}
A' = Q K^T + B = Q K^T + \alpha (Q \odot W_1) K^T,
\label{eq: attention score}
\end{equation}
where $\alpha = e^\eta$, $\eta$ is a learnable scaling factor and $\odot$ is the element-wise multiplication with channel-wise broadcasting.
Exploiting the distributive property of matrix multiplication over addition, Eq.~\ref{eq: attention score} can be rewritten as
\begin{equation}
A' = \left( Q + \alpha (Q \odot W_1) \right) K^T = Q'K^T,
\label{bias}
\end{equation}
where $Q' = Q \odot (\mathbf{1} + \alpha W_1)$, $\mathbf{1}$ is an all-ones matrix of the same dimension as $W_1$. 
This equation implies that the introduced bias term is equivalent to modulate the query matrix through $\mathbf{1} + \alpha W_1$.
Then the final attention weight $A = (a_{ij}) \in \mathbb{R}^{N \times N}$ is obtained by applying the softmax function to $A'$:
\begin{equation}
a_{ij} = \frac{\exp(\langle \tau_i q_i, k_j \rangle)}{\sum_{m=1}^N \exp(\langle \tau_i q_i, k_m \rangle)} 
= \frac{[\exp(\langle q_i, k_j \rangle)]^{\tau_i}}{\sum_{m=1}^N [\exp(\langle q_i, k_m \rangle)]^{\tau_i}},
\label{eq: attention temperature}
\end{equation}
where $q_i \in \mathbb{R}^{C_d}$, $k_j \in \mathbb{R}^{C_d}$ denote the $i$-th query and $j$-th key vectors, respectively. $\tau_i = 1 + \alpha w_i^{(1)}$ denotes the temperature value corresponding to the $i$-th query vector. Here $W_1$ is decomposed into $W_1 = [w_1^{(1)}, \dots, w_N^{(1)}]^T$, where $w_i^{(1)}$ is a scalar.
Notice that $\alpha > 0$, $w_i^{(1)}\in(0,1)$, the above attention weight $a_{ij}(\tau_i)$ allows controllable sharpening at queries with high matching confidence $(\tau_i >> 1)$, and recovers the standard softmax when queries are non-matchable $(\tau_i \approx 1)$. Theoretically, as the temperature increases ($\tau_i \to \infty$), the $a_{ij}$ converges to
\begin{equation}
\lim_{\tau_i \to \infty} a_{ij} = 
\begin{cases}
\frac{1}{|\mathcal{Z}_i|}, & j \in \mathcal{Z}_i \\
0, & \text{otherwise}
\end{cases}
\label{limit}
,\end{equation}
where $\mathcal{Z}_i = \left\{ j \mid \langle q_i, k_j \rangle = \max_m \langle q_i, k_m \rangle \right\}$ is the set of keys with maximum similarity to the query $q_i$. This limit corresponds to a differentiable approximation of the argmax operation, enabling the attention mechanism to focus sharply on the most relevant pixels while remaining trainable. In the case of multiple keys sharing the maximum score, the attention is distributed evenly among them, ensuring stability and avoiding ambiguity in the selection.

\noindent\textbf{Value Rescaling}. In addition to introducing the bias term before the softmax, we further adjust the attention weight after the softmax using the estimated confidence map $W_2 = [w_1^{(2)}, \dots, w_N^{(2)}]^T \in \mathbb{R}^{N \times 1}$. This adjustment is equivalent to an element-wise multiplication of the value matrix,
\begin{equation}
\label{eq:matchability_matrix_with_W2_subscript}
m_i = \sum_{j=1}^n a_{ij} \cdot w^{(2)}_j \cdot v_j = \sum_{j=1}^n a_{ij} \cdot \left( W_2 \odot V \right)_j
\end{equation}
where $v_j$ is the $j$-th value vector and $m_i \in \mathbb{R}^{C_d} $ is the $i$-th retrieved message corresponding to $q_i$. $\odot$ has the same meaning as in Eq.~\ref{bias}. To obtain the final updated query matrix $Q_{out}$, the up-sampling layer and the feed-forward network (FFN) are employed, 
\begin{equation}
Q_{out} = \text{FFN}(\text{Up}(\text{Re}(Q + M))),
\end{equation}
where $M$ is the message matrix composed of the vectors $m_i$. Re($\cdot$) reshapes tokens to 2D features, $\text{Up}(\cdot)$ is the up-sampling layer, and $\text{FFN}(\cdot)$ is the feed-forward network. This procedure is interleaved by $\text{T}$ times for obtaining the coarse-level feature descriptors $D^L_1, D^L_2\in\mathbb{R}^{H_cW_c \times C_d}$.

\subsection{Matching}
\textbf{Coarse Matching}.
The coarse similarity matrix is computed via scaled dot product,
\begin{equation}
S_{c}(i,j) = \frac{1}{\lambda} \left\langle D^L_1(i), D^L_2(j) \right\rangle,
\end{equation}
where $\lambda$ is a hyperparameter. Then we employ the row-wise and column-wise softmax operations to yield the probability matrix $P_c$,
\begin{equation}
P_c = \mathrm{softmax}_{row}(S_c) \cdot \mathrm{softmax}_{col}(S_c).
\end{equation}
Based on the probability matrix $P_c$, we select matches with confidence higher than a threshold $\theta_c$, and further enforce the mutual nearest neighbor (MNN) criterion. The coarse-level match prediction is denoted as
\begin{equation}
    \begin{aligned}
        \mathcal{M}_c = &\left\{ (i, j) \mid \forall (i,j)  \in \text{MNN}(P_c),\; P_c(i,j) \geq \theta_c \right\},
    \end{aligned}
    \label{eq:coarse-matches}
\end{equation}
where MNN($\cdot$) denotes the mutual nearest neighbor searching, which could filter out potential outlier matches.

\noindent \textbf{Fine Matching}. A two-stage refinement strategy is proposed to refine coarse-level matches to be fine-level matches. First, a lightweight feature fusion network is leveraged to fuse fine features $f^H_1, f^H_2$ and coarse-level feature descriptors $D^L_1, D^L_2$.
\begin{equation}
    \begin{aligned}
        D^H_k = \text{FN}(f^H_k, D^L_k),
    \end{aligned}
    \label{eq:feature fusion}
\end{equation}
where $k = 1, 2$, FN($\cdot$) denotes the feature fusion network and $D^H_k$ denotes the fused fine features.

After the fusion, at the first stage of the refinement, for each coarse-level match, two $8 \times 8$ local patches are extracted from $D^H_1, D^H_2$ to form $D^m_{1}, D^m_{2} \in \mathbb{R}^{N_m \times 64 \times C_f}$, where $N_m$ is the number of coarse matches. The fine similarity matrix $S_f \in \mathbb{R}^{N_m \times 64 \times 64}$ is computed via dot product between $D^m_{1}, D^m_{2}$. Then the dual-softmax is employed:
\begin{equation}
P_f = \mathrm{softmax}_{row}(S_f) \cdot \mathrm{softmax}_{col}(S_f),
\end{equation}
where $P_f$ is the fine probability matrix, which is also followed by MNN to produce one-to-one intermediate-level matches $\mathcal{M}_l = \{(i', j')\}$. At the second stage of the refinement, for each $i'$ in $I_1$, we perform the feature correlation within a $3 \times 3$ window centered at $j'$ in $I_2$. 
A softmax is then applied to obtain a match distribution, and the fine-level matches $\mathcal{M}_f = \{(i', j'')\}$ are computed via expectation over the distribution within the local window.

\subsection{Loss Function}
The overall loss function consists of four components: the coarse-level matching loss $\mathcal{L}_{c}$, the fine-level matching loss $\mathcal{L}_{f}$, $\mathcal{L}_{s}$, and the classification loss $\mathcal{L}_{m}$. Ground-truth correspondence matrices $P_{c}^{gt}$ and $P_{f}^{gt}$ are derived from known camera poses and depth information.

\noindent\textbf{Coarse-Level Matching Supervision}. The coarse-level loss $\mathcal{L}_{c}$ is formulated as the focal loss \cite{focal_loss} between the predicted $P_c$ and the coarse ground truth $P_c^{gt}$:
\begin{equation}
    \mathcal{L}_{c} = \mathrm{FL}(P_{c}^{gt}, P_c),
    \label{eq:coarse-loss}
\end{equation}
where $\mathrm{FL}(\cdot)$ denotes the focal loss function.

\noindent\textbf{Fine-Level Matching Supervision}. For the first stage of refinement, the $\mathcal{L}_{f}$ similarly adopts the focal loss \cite{focal_loss} to supervise the predicted fine probability matrix $P_f$,
\begin{equation}
    \mathcal{L}_{f} = \mathrm{FL}(P_{f}^{gt}, P_{f}).
    \label{eq:fine-loss}
\end{equation}
For the second refinement stage, we supervise the predicted coordinates $\mathbf{x}_i \in \mathbb{R}^2$, which represent the continuous 2D location obtained by mapping the predicted $j''$ to image coordinates.

The supervision is performed by minimizing a masked L2 loss $\mathcal{L}_s$ with respect to the corresponding ground-truth $\mathbf{x}^{\text{gt}}_i \in \mathbb{R}^2$,
\begin{equation}
\mathcal{L}_s = 
\mathbb{E}_{i \sim \mathcal{U}(\mathcal{G})} \left[ 
\left\| \mathbf{x}_i - \mathbf{x}_i^{\text{gt}} \right\|_2^2 
\right],
\label{eq:local_loss_expectation}
\end{equation}
where $\mathcal{G} = \left\{ i \,\middle|\, \left\| \mathbf{x}_i^{\text{gt}} \right\|_{\infty} < \epsilon \right\}$, $\mathcal{U}(\mathcal{G})$ denotes the uniform distribution over $\mathcal{G}$, and $\epsilon$ is a constant.

\noindent\textbf{Matching Confidence Supervision}. For better supervising the estimation of confidence maps, a classification loss $\mathcal{L}_m$ is proposed. Given the ground-truth $P_c^{gt} \in \{0,1\}^{H_cW_c \times H_cW_c}$, we compute two ground-truth confidence maps by summing over each view dimension,
\begin{equation}
    P^{gt}_{c,1} = \sum_j P^{gt}_c(i,j), \quad 
    P^{gt}_{c,2} = \sum_i P^{gt}_c(i,j),
\end{equation}
where $P^{gt}_{c,1}$ and $P^{gt}_{c,2}$ denote the ground-truth confidence maps for the first and second image, respectively.
The predicted confidence maps $\hat{W}_1$ and $\hat{W}_2$ are supervised through the binary cross-entropy loss BCE($\cdot$),
\begin{equation}
\mathcal{L}_{\text{m}_1} = \mathrm{BCE}(\hat{W}_1, P^{gt}_{c,1}),
\mathcal{L}_{\text{m}_2} = \mathrm{BCE}(\hat{W}_2, P^{gt}_{c,2}).
\end{equation}
The final classification loss is given by $\mathcal{L}_m = (\mathcal{L}_{\text{m}_1}+\mathcal{L}_{\text{m}_2}) / 2 $.
The overall loss function $\mathcal{L}$ is the combination of the four components, as follows:
\begin{equation}
    \mathcal{L} = \mathcal{L}_c + \mathcal{L}_f + \mathcal{L}_s + \beta \mathcal{L}_m,
\end{equation}
where $\beta$ is a hyperparameter that can be tuned to balance the contribution of the classification loss $\mathcal{L}_m$.

\begin{table*}[t]
    \centering
    \resizebox{\linewidth}{!}{
    \begin{tabular}{clccccccc} 
    \toprule
    \multirow{2}{*}{Category} & \multirow{2}{*}{Method} & \multicolumn{3}{c}{ScanNet}  & \multicolumn{3}{c}{MegaDepth}   \\ 
    \cmidrule(lr){3-5}
    \cmidrule(lr){6-8}
        &     &AUC@5$^\circ$ & AUC@10$^\circ$ & AUC@20$^\circ$     &AUC@5$^\circ$       &AUC@10$^\circ$       &AUC@20$^\circ$     \\ 
    \midrule
    \multirow{4}{*}{Sparse} 
    & SP \cite{SuperPoint}  + SG \cite{Superglue} $_{\textcolor{gray}{\textit{CVPR'20}}}$ & 16.2 &32.8 & 49.7& 57.6 &72.6 &83.5 \\
    & SP \cite{SuperPoint}  + LG \cite{lightglue} $_{\textcolor{gray}{\textit{ICCV'23}}}$& 14.8 & 30.8 & 47.5 & 58.8 &73.6 &84.1  \\
    & XFeat \cite{xfeat} $_{\textcolor{gray}{\textit{CVPR'24}}}$  & 16.7  & 32.6 & 47.8 & 44.2 &58.2 &69.2   \\
    & DeDoDe$_B$ \cite{dedode} $_{\textcolor{gray}{\textit{3DV'24}}}$ &- &- &- &61.1 &73.8 &83.0   \\

    \hline
    \multirow{7}{*}{Semi-Dense} 
    & LoFTR \cite{LoFTR} $_{\textcolor{gray}{\textit{CVPR'21}}}$ & 16.9 & 33.6 & 50.6 &62.1&75.5&84.9  \\
    & MatchFormer \cite{matchformer} $_{\textcolor{gray}{\textit{ACCV'22}}}$ &15.8 & 32.0 & 48.0 &62.0&75.6&84.9  \\
    & ASpanFormer \cite{aspanformer} $_{\textcolor{gray}{\textit{ECCV'22}}}$ &\cellcolor{third}19.6 &\cellcolor{third} 37.7 & \cellcolor{third}54.4 & 62.6&76.1&85.7  \\
    &XFeat \cite{xfeat} $^\star$ $_{\textcolor{gray}{\textit{CVPR'24}}}$ & 18.4& 34.7& 50.3 & 50.8 &66.8 &78.8   \\
    & RCM \cite{raise_ceiling} $_{\textcolor{gray}{\textit{ECCV'24}}}$ & 17.3 & 34.6 & 52.1 & 58.3 & 72.8 & 83.5  \\
    & ELoFTR \cite{Eloftr} $_{\textcolor{gray}{\textit{CVPR'24}}}$ &19.2 & 37.0& 53.6 &63.7&77.0&86.4  \\
    & JamMa \cite{jamma} $_{\textcolor{gray}{\textit{CVPR'25}}}$ & 14.7 & 30.2 & 46.5 &\cellcolor{third} 64.1&\cellcolor{third} 77.4& \cellcolor{third}86.5  \\
    & CoMatch \cite{comatcher} $_{\textcolor{gray}{\textit{ICCV'25}}}$ & \cellcolor{second}21.7 & \cellcolor{second}40.2 & \cellcolor{second}56.7 & \cellcolor{second}65.5 & \cellcolor{second} 78.2 & \cellcolor{second}87.1  \\
    & Proposal &\cellcolor{first}21.9 &\cellcolor{first} 40.4 & \cellcolor{first}57.1 &\cellcolor{first} 66.0& \cellcolor{first}78.9&\cellcolor{first} 87.9  \\
    \bottomrule
    \end{tabular}
    }
    \caption{\textbf{Results of Relative Pose Estimation on MegaDepth and ScanNet Datasets.}
    Pose error AUCs at three thresholds are reported.
    Lo-RANSAC is employed when evaluated on MegaDepth, while RANSAC is employed when evaluated on ScanNet.
    The \colorbox{first}{1st}, \colorbox{second}{2nd}, and \colorbox{third}{3rd}-best methods are highlighted.
    }
    \label{tab:relativepose}
    \vspace{-0.3in}
    \end{table*}

\section{Experiments}
The proposed method is evaluated on relative pose estimation, image matching and visual localization tasks. Efficiency analysis confirms its practicality, and ablation studies provide insights into the contributions of each component.

\subsection{Relative Pose Estimation}
\textbf{Datasets.} In this section, our method is trained on MegaDepth \cite{Megadepth} and evaluated on the outdoor dataset MegaDepth and the indoor dataset ScanNet \cite{Scannet}. Images of MegaDepth are resized as $832 \times 832$ pixels while images of ScanNet are resized as $640 \times 480$ pixels. When evaluated on ScanNet, we choose the outdoor version of all baselines for a fair comparison.

\noindent\textbf{Metrics.} The metric used in outdoor pose estimation is the AUC of pose errors under different thresholds $\left\{5^\circ, 10^\circ, 20^\circ\right\}$. The pose error is defined as the maximum of the angular error in rotation and translation. The match is considered correct if its epipolar error is below 5e-4. The Lo-RANSAC is performed on MegaDepth, while RANSAC is performed on ScanNet. The RANSAC threshold of all methods is set to 0.5. 

\noindent\textbf{Results.} In addition to semi-dense methods \cite{xfeat, jamma, raise_ceiling}, we also compare our method against sparse methods \cite{Superglue, lightglue, dedode}. Table~\ref{tab:relativepose} demonstrates that our method outperforms both sparse and semi-dense methods by a significant margin. We attribute the prominence to the proposed confidence-guided attention, which helps reduce the interaction with non-matchable regions. To mitigate the randomness introduced by RANSAC and LO-RANSAC, we first shuffle the set of matches generated by each matcher before performing the relative pose estimation.
\vspace{-0.1in}
\subsection{Image Matching}
\begin{figure*}[t]
\centering
\includegraphics[width=\columnwidth]{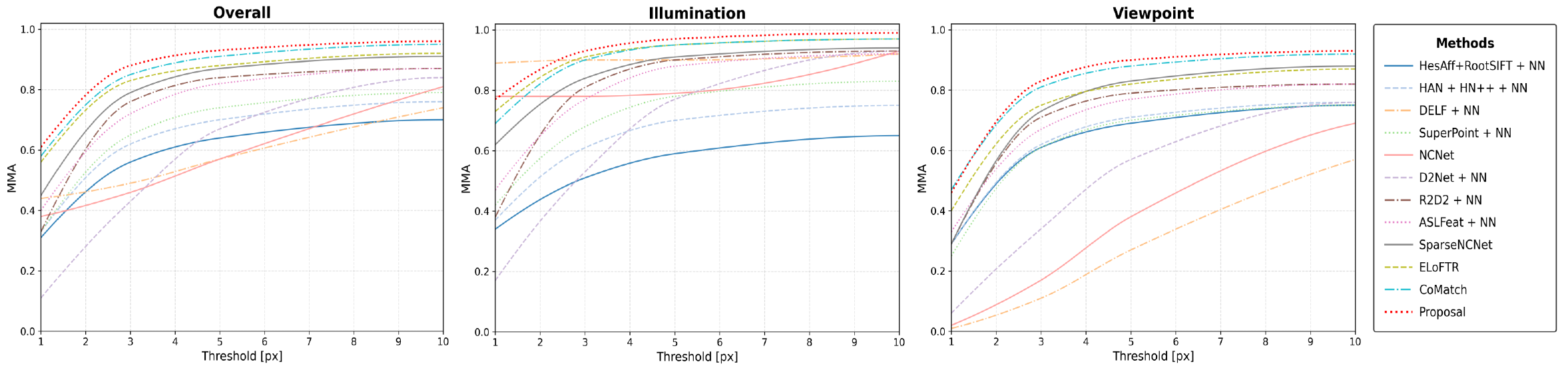} 
\vspace{-0.3in}
\caption{\textbf{Image matching results on HPatches.} MMA curves are plotted by varying the reprojection error threshold.}
\label{matching}
\vspace{-0.3in}
\end{figure*} 
\textbf{Datasets.} Since the RANSAC could bring randomness in the relative pose estimation, to better understand the performance of matchers, we use the mean matching accuracy (MMA) to evaluate the ability to extract and match local features. In this section, all methods are evaluated on the HPatches \cite{HPatches} dataset. 

\noindent\textbf{Metrics.} We follow the setup proposed in D2Net \cite{D2Net} and report the mean matching accuracy (MMA) \cite{MMA} under pixel thresholds $\left\{1, 3, 5, 10\right\}$.

\noindent\textbf{Results.}
In addition to the aforementioned baselines, several other methods are included for comparison in the image matching task, namely: HesAff \cite{HesAff}, HAN \cite{HAN}, HN++ \cite{HN++}, DELF \cite{DELF}, D2Net \cite{D2Net}, R2D2 \cite{R2D2}, SparseNCNet \cite{ENCNet}, NCNet \cite{NCNet}.
As shown in Fig.~\ref{matching}, our method outperforms all baselines. Notably, these gains in precise pixel-level matching are not fully captured by pose estimation metrics such as AUC, highlighting the additional value of our approach for tasks that require exact correspondence accuracy.
\vspace{-0.2in}
\subsection{Visual Localization}

\noindent\textbf{Datasets}. Visual localization is an important downstream task of image matching, aiming to estimate the 6-DoF poses of query images with respect to a pre-built 3D scene model. We evaluate our method on the Aachen v1.1 dataset \cite{achaenv1.1}, a widely used large-scale outdoor benchmark that features significant viewpoint variation and challenging day–night illumination changes.

We follow the standard evaluation protocol by adopting the open-source HLoc framework \cite{hloc}, as in \cite{Eloftr}. The evaluation reports the percentage of localized queries whose predicted poses fall within specified thresholds of position and orientation errors. Results are reported separately for daytime and nighttime queries using the full localization track of the benchmark.

\begin{table}[t]
\centering
\label{tab:visual_localization}
\setlength{\tabcolsep}{12pt}
\resizebox{0.8\columnwidth}{!}{
\begin{tabular}{lcc}
\toprule
\textbf{Method} & \textbf{Day} & \textbf{Night} \\
& \multicolumn{2}{c}{(0.25m, 2°) / (0.5m, 5°) / (1m, 10°)} \\
\midrule
DeDoDe\textsubscript{B}$_{\textcolor{gray}{\textit{3DV'24}}}$ & 87.4 / 94.7 / 98.5 & 70.7 / 88.0 / 97.9 \\
SP + LG$_{\textcolor{gray}{\textit{ICCV'23}}}$ & \textbf{89.6} / \textbf{95.8} / \textbf{99.2} & 72.8 / 88.0 / 99.0 \\
LoFTR*$_{\textcolor{gray}{\textit{CVPR'21}}}$ & 89.1 / 95.8 / 98.8	& 76.4 / 90.1 / 99.0\\
ASpanFormer$_{\textcolor{gray}{\textit{ECCV'22}}}$ & 89.4 / 95.6 / 99.0 & \textbf{77.5} / \textbf{91.6} / 99.0 \\
ELoFTR*$_{\textcolor{gray}{\textit{CVPR'24}}}$ & 87.5 / 95.1 / 98.3 & 76.4 / 89.0 / 99.0 \\
JamMa$_{\textcolor{gray}{\textit{CVPR'25}}}$ & 87.7 / 95.1 / 98.4 & 73.3 / \textbf{91.6} / 99.0 \\
% CoMatch$_{\textcolor{gray}{\textit{ICCV'25}}}$ & - / - / - & - / - / - \\
Proposal & 88.0 / 95.3 / 98.7 & 77.0 / 90.6 / \textbf{99.5} \\
\bottomrule
\end{tabular}
}
\caption{\textbf{Visual Localization on the Aachen Day-Night Benchmark v1.1.} * denotes the reproduced results under the same settings with ours.}
\vspace{-0.45in}
\label{vloc}
\end{table}

\noindent \textbf{Metric}. We report the percentage of query images that are successfully localized within three increasingly relaxed thresholds of position and orientation error. A predicted camera pose is considered correct under a given threshold if the estimated translation error is less than the specified distance (e.g., 0.25m) and the rotation error is less than the specified angular deviation (e.g., 2°) from the ground truth. This joint criterion ensures that both spatial and directional accuracy are taken into account.

\noindent \textbf{Results}. Table \ref{vloc} compares our proposed method with other state-of-the-art (SOTA) matching approaches. \textbf{Fair evaluation in visual localization remains challenging, primarily due to the lack of open-source settings and incomplete reporting of key experimental details in many existing methods. Notably, localization performance is highly sensitive to parameters such as the quantization patch size and the maximum permissible keypoint assignment error.} To ensure a rigorous comparison, all methods are evaluated under a unified setting. In our experiments, we use a patch size of 8 and a keypoint error threshold of 4 pixels. Results marked with * denote reproduced outcomes with our unified settings.
\vspace{-0.15in}
 
\begin{table}[t]
\centering

% ================= Left: Original Efficiency Table ================= %
\begin{minipage}[b]{0.49\textwidth}
\centering
\setlength{\tabcolsep}{6pt}
\resizebox{0.99\linewidth}{!}{
\begin{tabular}{clcc}
    \toprule
    Category & Method & Params (M) $\downarrow$ & Time (ms) $\downarrow$ \\
    \midrule
    \multirow{2}{*}{Sparse} 
    & SP + SG & 13.3 & 96.9 \\
    & SP + LG & \cellcolor{third}13.2 & \cellcolor{third}84.2 \\
    \midrule
    \multirow{2}{*}{Dense}
    & DKM & 72.3 & 554.2 \\
    & RoMa & 111.3 & 824.9 \\
    \midrule
    \multirow{5}{*}{Semi-Dense}
    & LoFTR & \cellcolor{first}11.6 & 117.5 \\
    & MatchFormer & 20.3 & 186.0 \\
    & AspanFormer & 15.8 & 155.7 \\
    & ELoFTR & 16.0 & \cellcolor{first}69.6 \\
    & CoMatch & \cellcolor{second}12.0 & 99.2 \\
    & Proposal & 16.0 & \cellcolor{second}73.4 \\
    \bottomrule
\end{tabular}
}
\captionof{table}{The efficiency comparison of different categories of methods.}
\label{efficiency}
\end{minipage}
\vspace{-0.2in}
\hfill
% ================= Right: Your Component Figure ================= %
\begin{minipage}[b]{0.5\textwidth}
\centering
\includegraphics[width=\linewidth]{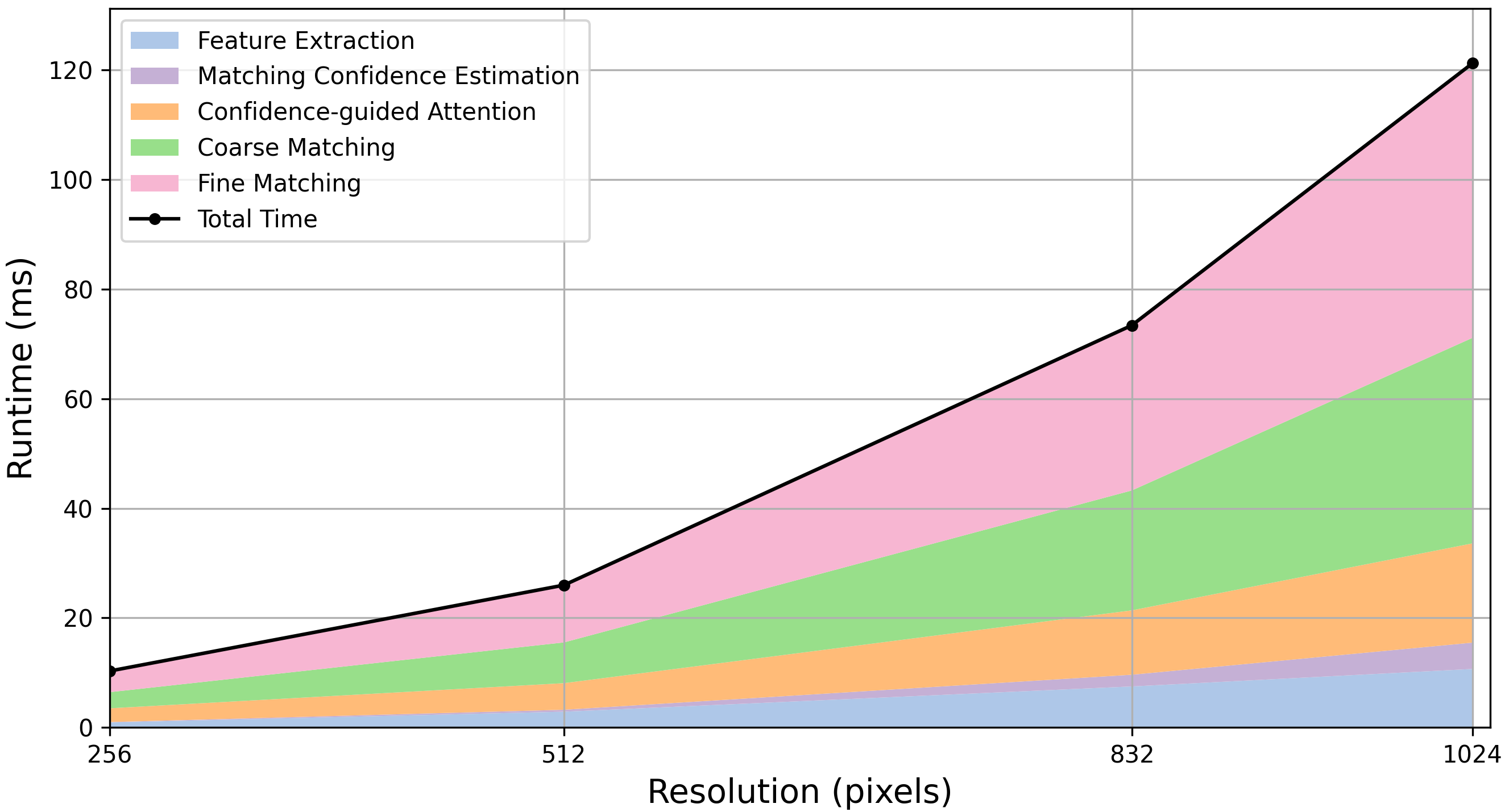}
\captionof{figure}{Component analysis of runtime.}
\label{component}
\end{minipage}
\vspace{-0.2in}
\end{table}

\subsection{Efficiency Analysis.}
\textbf{Efficiency Comparison.} In this section, we compare the efficiency of various matching methods. Experimental settings are the same as in the outdoor pose estimation on MegaDepth. As shown in Table~\ref{efficiency}, dense methods like DKM~\cite{dkm} and RoMa~\cite{RoMa} incur significantly higher computational costs. In contrast, our proposed method achieves a better trade-off between performance and efficiency, with fewer parameters and faster inference. This makes it more suitable for time-sensitive or resource-limited applications.

\noindent\textbf{Component Analysis of Runtime.} Fig. \ref{component} illustrates the runtime breakdown of each module across varying image resolutions. The Matching Confidence Estimation module exhibits the lowest computational overhead, primarily due to the absence of additional learnable parameters. In contrast, both Coarse Matching and Fine Matching incur substantially higher runtime, which scales rapidly with increasing image resolution. This trend is largely attributed to the growth in the number of detected coarse-level matched points. Therefore, reducing the computational cost of the matching stage represents a promising direction for future research in local feature matching.
\vspace{-0.2in}
\subsection{Ablation Studies}
In the ablation studies, results on ScanNet are reported, following the same experimental protocol as that used in the experiment of indoor pose estimation.

\noindent\textbf{Ablation on the effectiveness of proposed modules.} In this section, we conduct a detailed analysis of the effectiveness of each proposed module. In Table \ref{ablation_on_main_designs}, we progressively introduce the Confidence-guide Bias, Value Rescaling, and Matching Confidence Supervision modules. Experimental results demonstrate that each proposed module contributes to improving matching accuracy. As the Matching Confidence Supervision is meaningful only when confidence maps are incorporated into attention, its effectiveness is evaluated jointly with another two modules, rather than individually.
\begin{figure*}[t]
\centering
\includegraphics[width=\textwidth]{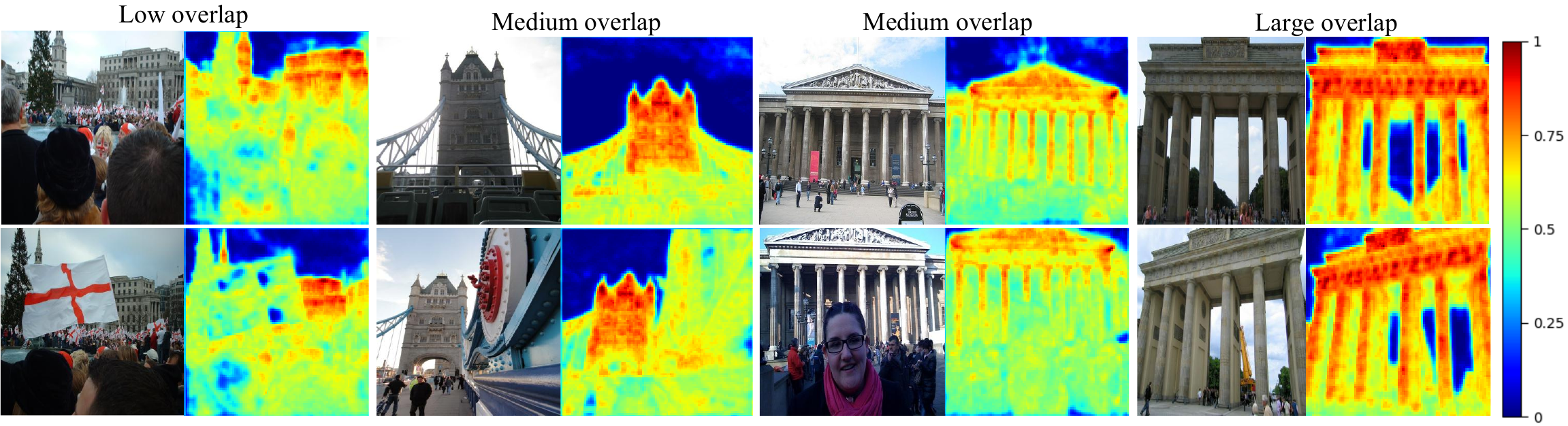} 
\vspace{-0.3in}
\caption{The visualization results of confidence maps.}
\label{visualization}
\vspace{-0.2in}
\end{figure*}  

\begin{table}[t]
\centering
\captionsetup{skip=3pt}

% ---------- Left ----------
\begin{minipage}[t]{0.48\textwidth}
\centering
\setlength{\tabcolsep}{3pt}

\resizebox{\linewidth}{!}{
\begin{tabular}{ccccccc}
    \hline
    \multirow{2}{*}{\makecell{Conf.-\\guided Bias}} & 
    \multirow{2}{*}{\makecell{Value \\ Rescaling}} & 
    \multirow{2}{*}{\makecell{Matching Conf. \\ Supervision}} & 
    \multicolumn{3}{c}{Pose estimation AUC} & 
    \multirow{2}{*}{Prec} \\
    & & & @$5^\circ$ & @$10^\circ$ & @$20^\circ$ & \\
    \hline
    \xmark & \xmark & \xmark & 19.23 & 37.01 & 53.58 & 70.08 \\
    \cmark & \xmark & \xmark & 20.11 & 38.42 & 55.33 & 71.44\\
    \xmark & \cmark & \xmark & 19.88 & 37.75 & 54.06 & 70.52\\
    \cmark & \cmark & \xmark & \cellcolor{second}21.21 & \cellcolor{second}39.74 & \cellcolor{second}56.65 & \cellcolor{second}71.86\\
    \xmark & \xmark & \cmark & 20.64 & 39.11 & 56.03 & 71.83\\
    \cmark & \cmark & \cmark & \cellcolor{first}21.94 & \cellcolor{first}40.42 & \cellcolor{first}57.10 & \cellcolor{first}72.01\\
    \hline
\end{tabular}
}
\caption{Ablation study on the effectiveness of proposed modules.}
\label{ablation_on_main_designs}
\vspace{-0.5in}
\end{minipage}
\hfill
% ---------- Right ----------
\begin{minipage}[t]{0.48\textwidth}
\centering
\setlength{\tabcolsep}{3pt}

\resizebox{\linewidth}{!}{
\begin{tabular}{lcccc}
    \hline
    \multirow{2}{*}{Methods} & 
    \multicolumn{3}{c}{Pose estimation AUC} & 
    \multirow{2}{*}{Prec} \\
    & @$5^\circ$ & @$10^\circ$ & @$20^\circ$ & \\
    \midrule
    i) $\hat{W}_i = \sigma(\tilde{W}_i)$ & 20.04 & 39.78 & 56.13 & 71.61\\
    ii) $\hat{W}_i = \text{ReLU}(\tilde{W}_i - \text{mean}(\tilde{W}_i))$ & 20.24 & 39.96 & 56.26 & 71.40\\
    iii) $\hat{W}_i = \sigma(\tilde{W}_i - \text{mean}_{row/col}(\tilde{W}_i))$ & 
    \cellcolor{second}21.43 & \cellcolor{second}40.12 & \cellcolor{second}56.74 & \cellcolor{third}71.77\\
    iv) $\hat{W}_i = \sigma(\text{conv}_{1\times1}(\tilde{W}_i))$ & 
    \cellcolor{third}21.21 & \cellcolor{third}40.03 & \cellcolor{third}56.65 & \cellcolor{second}71.94\\
    v) $\hat{W}_i = \sigma(\tilde{W}_i - \text{mean}(\tilde{W}_i))$ & 
    \cellcolor{first}21.94 & \cellcolor{first}40.42 & \cellcolor{first}57.10 & \cellcolor{first}72.01\\
    \hline
\end{tabular}
}
\caption{Ablation study on the confidence maps, $i = 1, 2$.}
\label{ablation_on_matchability_map}
\vspace{-0.5in}
\end{minipage}
\end{table}

\noindent\textbf{Ablation on confidence maps.} As shown in Table \ref{ablation_on_matchability_map}, in addition to the proposed method, we further investigate four alternative approaches for obtaining confidence maps. $\sigma(\cdot)$ denotes the sigmoid function. The design of i) is too straightforward and can not achieve the optimal performance. When the ReLU function is applied in ii), non-matchable pixels are suppressed to zero, leading to sparse attention. This sparsity hinders the aggregation of global contexts and results in less discriminative features.

Similar to detector-based methods, iii) could only preserve the salient regions and narrow the potential matching space. Compared to iii), in v), we leverage the global mean values of $\tilde{W}_1$, $\tilde{W}_2$ rather than row- or column-wise mean values $\mathrm{mean}_{row}(S)$ or $ \mathrm{mean}_{col}(S)$. This design avoids over-suppressing low-texture regions, which often carry important structural information in indoor scenes.

With respect to iv), we adopt a learned approach to estimate the confidence maps. However, experimental results show a slight performance drop. This is because the proposed method is trained on the outdoor dataset MegaDepth \cite{Megadepth} rather than the indoor dataset ScanNet \cite{Scannet}. This could raise concerns about the generalization of the learned confidence maps. In contrast to the learned option, we leverage v) to incorporate a heuristic prior derived from the mutual similarity of CNN's features, which can help the model handle unseen cases.
% \begin{figure*}[t]
% \centering
% \includegraphics[width=\textwidth]{Visio-qualitative.pdf} % Reduce the figure size so that it is slightly narrower than the column.
% \vspace{-0.2in}
% \caption{Qualitative comparison with existing methods on MegaDepth. The red color indicates epipolar error beyond $5 \times 10^{-4}$.}
% \vspace{-0.2in}
% \label{qualtative}
% \end{figure*}

\noindent\textbf{Discussion about the robustness of confidence maps.} For a complete analysis, visualization results on pairs with different overlaps are shown in Fig. \ref{visualization}. A brighter color indicates a higher probability of a region being matchable. Confidence maps serve as informative priors that narrow the matching search space by excluding irrelevant regions. Although the confidence maps are not always perfectly accurate, they generally highlight high-confidence regions around stable structures such as distinctive textures. When the confidence maps become less reliable due to appearance changes or partial occlusions, the learnable parameter $\alpha$ (Eq.~\ref{bias}) adaptively modulates the attention weights, allowing the network to down-weight uncertain pixels and maintain robust matching.
\begin{table}[t]
\centering
\captionsetup{skip=3pt}

% -------- Left Table --------
\begin{minipage}[t]{0.48\textwidth}
\centering
\setlength{\tabcolsep}{3pt}

\resizebox{\linewidth}{!}{
\begin{tabular}{lcccc}
    \hline
    \multirow{2}{*}{Methods} & \multicolumn{3}{c}{Pose estimation AUC} & \multirow{2}{*}{Prec} \\
                            & @$5^\circ$ & @$10^\circ$ & @$20^\circ$ & \\
    \midrule
    i) $B_1 = \alpha (Q \odot W_1)K^T$ & \cellcolor{first}21.94 & \cellcolor{first}40.42 & \cellcolor{first}57.10 & \cellcolor{first}72.01\\
    ii) $B_2 = \alpha Q (K\odot W_2)^T$ & \cellcolor{second}20.45 & 39.80 & 56.69 & \cellcolor{second}71.53\\
    iii) $B_3 = \alpha (Q \odot W_1) (K\odot W_2)^T$& 20.30 & \cellcolor{second}39.92 & \cellcolor{second}56.81 & 71.37\\
    \hline
\end{tabular}
}

\caption{Ablation study on the confidence-guided bias.}
\label{ablation_on_bias}
\vspace{-0.5in}
\end{minipage}
\hfill
% -------- Right Table --------
\begin{minipage}[t]{0.48\textwidth}
\centering
\setlength{\tabcolsep}{11pt}

\resizebox{\linewidth}{!}{
\begin{tabular}{lcccc}
    \hline
    \multirow{2}{*}{The value of $\beta$} & \multicolumn{3}{c}{Pose estimation AUC} & \multirow{2}{*}{Prec} \\
                            & @$5^\circ$ & @$10^\circ$ & @$20^\circ$ & \\
    \hline
    i) 0.2 & 21.52 & 40.04 & 56.93 & 71.93\\
    ii) 0.5 & \cellcolor{second}21.73 & \cellcolor{second}40.27 & \cellcolor{second}57.01 & \cellcolor{second}71.99\\
    iii) 1.0 & \cellcolor{first}21.94 & \cellcolor{first}40.42 & \cellcolor{first}57.10 & \cellcolor{first}72.01\\
    \hline
\end{tabular}
}
\caption{Ablation study on the contribution of $\mathcal{L}_m$.}
\label{ablation_on_Lc}
\end{minipage}
\vspace{-0.5in}
\end{table} 

\noindent\textbf{Ablation on the confidence-guided bias.} In this section, we analyze the three variants of the confidence-guided bias $B_1$, $B_2$, and $B_3$ on ScanNet. As shown in Table \ref{ablation_on_bias}, the $B_1$ could achieve the optimal performance at the AUC scores and matching precision. Interestingly, the overall performance could degrade when $W_2$ is introduced. This may be because we have already introduced the $W_2$ at the value matrix of the attention process, and the additional information in the attention bias could result in information redundancy, which hinders the precise localization of matching points.
\begin{figure*}[t]
\centering
\includegraphics[width=\columnwidth]{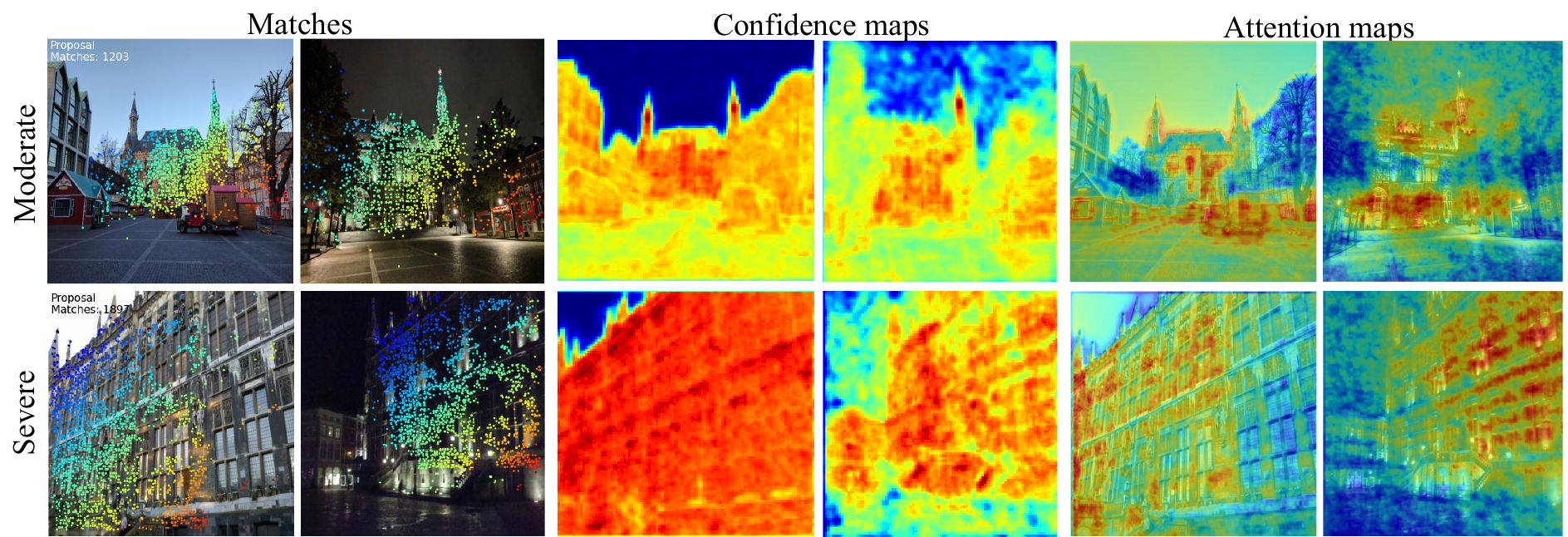} 
\vspace{-0.3in}
\caption{Sensitivity analysis on day-night image pairs.}
\label{sensitivity}
\vspace{-0.3in}
\end{figure*}
\noindent\textbf{Ablation on the contribution of $\mathcal{L}_m$.} The hyperparameter $\beta$ balances the contribution of the classification loss $\mathcal{L}_m$ with the overall training objective $\mathcal{L}$. As shown in Table \ref{ablation_on_Lc}, a small $\beta$ implies a weak contribution from $\mathcal{L}_m$, causing the confidence maps to rely mainly on the CNN backbone’s implicit feature correlations, which may not accurately reflect the true co-visibility relationship between pixels. As $\beta$ increases, $\mathcal{L}_m$ imposes stronger constraints derived from ground-truth co-visible labels computed from depth and relative poses, guiding the backbone to encode more geometrically consistent features. This leads to more reliable confidence estimation and improves the overall performance.

\noindent\textbf{Sensitivity analysis about imperfect confidence maps.}
We use day–night image pairs for the sensitivity analysis and categorize the cases according to the corruption level of confidence maps, as shown in Fig. \ref{sensitivity}.
In the extreme case where the confidence map assigns high confidence to nearly all pixels, the prior becomes non-informative but the sharpness of attention is jointly controlled by the confidence map and the learnable parameter $\alpha$.
When the attention is sufficiently sharper, the query pixel only focuses on most similar target pixels (Eq. \ref{limit}), making the model robust to noisy or corrupted confidence estimation.
\vspace{-0.2in}
\section{Conclusion}
In this work, we proposed a semi-dense matching method that addresses the inefficiency of uniformly treating all pixels in previous methods. By introducing a matching-confidence prior, our method adaptively reweights the attention distribution, leading to more robust and accurate correspondence estimation. Extensive experiments on various indoor and outdoor benchmarks demonstrate that our method consistently outperforms existing baselines.

% \section*{Acknowledgements}
% Please insert your acknowledgments here.

% ---- Bibliography ----
%
% BibTeX users should specify bibliography style 'splncs04'.
% References will then be sorted and formatted in the correct style.
%
\bibliographystyle{splncs04}
\bibliography{main}
\end{document}